# IMPROVE CAPTCHA'S SECURITY USING GAUSSIAN BLUR FILTER

Ariyan Zarei [1]

[1] Computer Science Student, Shahid Beheshti University, Tehran, Iran
Member of Young Researchers Club, Karaj branch, Islamic Azad University, Karaj, Iran
`arianzarei73@gmail.com`

## ABSTRACT

*Providing security for webservers against unwanted and automated registrations has become a big concern. To prevent these kinds of false registrations many websites use CAPTCHAs. Among all kinds of CAPTCHAs OCR-Based or visual CAPTCHAs are very common. Actually visual CAPTCHA is an image containing a sequence of characters. So far most of visual CAPTCHAs, in order to resist against OCR programs, use some common implementations such as wrapping the characters, random placement and rotations of characters, etc. In this paper we applied Gaussian Blur filter, which is an image transformation, to visual CAPTCHAs to reduce their readability by OCR programs. We concluded that this technique made CAPTCHAs almost unreadable for OCR programs but, their readability by human users still remained high.*

## 1. INTRODUCTION

Nowadays many people use internet to provide them with their needs such as shopping, banking transactions, registrations, communications, etc. Protecting servers and websites on internet from dangerous threats and attacks has become a serious problem. One of the important and dangerous threats is automated false registrations on webservers that can waste the recourses and finally cause serious damages to the servers.

Many of the registration servers use CAPTCHAs to prevent these attacks. CAPTCHA stands for "Completely Automated Public Turing Test to Tell Computers and Humans Apart" [1]. Actually CAPTCHA is a test that can distinguish human users from robots and programs. Common kinds of CAPTCHAs are audio CAPTCHAs and visual CAPTCHAs.

Most of the websites and registration systems use the visual CAPTCHAs which is usually an image containing sequence of characters with some noises. Websites show this images to users and ask them to enter characters correctly. Because of the weak points of OCR applications this kind of CAPTCHAs has become very common. However, the OCR applications have improved over time and many of them are able to remove the noises and recognize the words.

In this research we have applied Gaussian Blur filter on visual CAPTCHAs and we have investigated that how much they are secured against OCR programs and readable by humans.

## 2. PREVIOUS KNOWLEDGE

Generally there are various kinds of CAPTCHAs, OCR-Based CAPTCHAs like Gimpy method, Pattern recognition CAPTCHAs like BONGO, and Sound-Based CAPTCHAs.

The BONGO CAPTCHA asks the user to solve a visual pattern recognition problem. It shows 2 series of shapes and patterns with different colors and sizes, then shows another pattern and asks the user to determine that the shown pattern belongs to which of the two series of patterns. [1]

The Sound-Based CAPTCHAs was first designed by Nancy Chan in the University of Hong Kong. It is a system that plays a sound clip containing words and numbers and asks the user to enter what he/she heard. [1]

In this research we worked on the OCR-Based or Visual CAPTCHAs. The idea of visual CAPTCHAs was first created to prevent automated and futile registrations on AltaVista website in 1997. It was done by Andrei Broder and then by DEC Systems Research Center. After that in 2000 Yahoo company decided to have powerful and "easy to use Turing test" to prevent unwanted registrations on its services such as chat room and Email. This system designed by Prof. Manuel Blum at school of computer science at Carnegie Mellon University. [2]

This kind of Turing test first called CAPTHCA by Luis von Ahn, Manuel Blum, Nicholas Hopper, and John Langford from Carnegie Mellon University. [3]

So far visual CAPTCHAs have some common implementations like wrapping the characters via Linear Transformations, Random placement of characters, Noises applied to background, adding horizontal lines over the characters, etc. [4]

The webservers use different CAPTCHAs with different implementations and methods. For example the Gimpy method which is an OCR-Based CPTCHA works by creating an image with colored and noised background containing 7 words with wrapped characters, Then asking the user to enter 3 words out of the 7 words in the picture. [1]

We have other methods but most of them use the common implementations and techniques that we mentioned above.

## 3. METHODOLOGY

What we have done in this research was applying Gaussian Blur filter to some CAPTCHAs and test their readability by human users and OCR systems.

### 3.1. Gaussian Blur

Gaussian Blur is a kind of image filter that uses the Gaussian function to make an image blurred. The two dimensional Gaussian function is:

$$G(x,y) = \frac{1}{2\pi\sigma^2} e^{-\frac{x^2+y^2}{2\sigma^2}}$$

X is the distance from horizontal axis, y is the distance from vertical axis and $\sigma$ is the standard deviation of the Gaussian distribution which is also related to Radius of Blurriness in image processing apps.

In order to apply this filter to an image, we have to generate a surface with Gaussian function. The contours of the surface are concentric circles with Gaussian distribution from center point.

Values from this distribution build a convolution matrix which should be applied to image to reach final result. [5]

### 3.2. Convolution Matrix and the Way of Applying It to an Image

Actually Convolution Matrix is a small matrix with usually $3 \times 3$ or $5 \times 5$ dimension. It also called Kernel or Mask. It is used to apply some filters to images. For example figure 1 shows the Convolution Matrix of Edge Detection filter.

$$\begin{bmatrix} -1 & -1 & -1 \\ -1 & 8 & -1 \\ -1 & -1 & -1 \end{bmatrix}$$

Figure 1. Convolution Matrix of Edge Detection

To apply a filter to an image, each Kernel value must be multiplied by corresponding input image pixel value. The output pixel is the average (or sometimes the sum) of these values. Figure 2 determines the process.

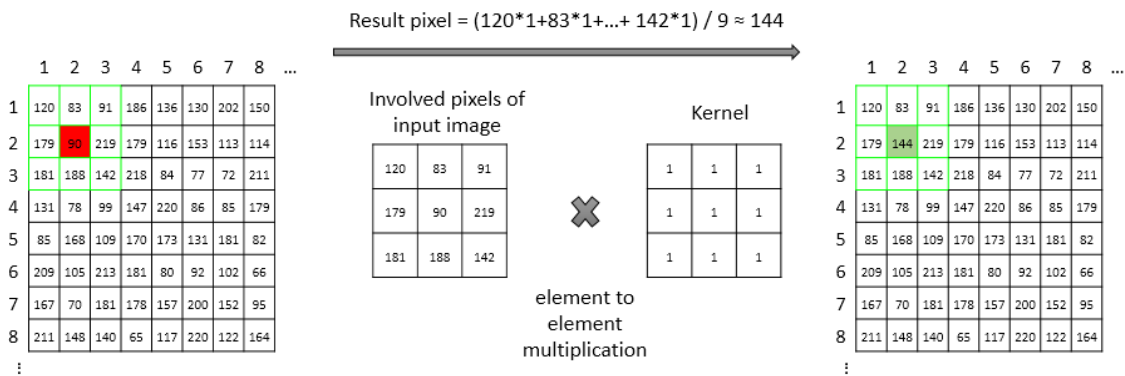

Figure 2. Applying Convolution Matrix to image

### 3.3. Applying Gaussian Blur to CAPTCHAs

We have generated 50 simple OCR-Based CAPTCHAs with simple white background and simple characters. These CAPTCHAs contained two meaningless 4-7 letters word separated by space. We created these images by an application that we've written in C# language. In this application we generate two random word as mentioned above and then we draw it to an image.

Then we applied Gaussian filter on these images with Radius of 1 and Radius of 2 by using another application in C# as you can see it in figure 4. In this application we surveyed on the image's pixels and applied the convolution matrix to each pixel. You can see some examples of these CAPTCHAs in figure 3.

Figure 3. Visual CAPTCHAs with Gaussian Blur filter applied on.
Left: Gaussian Blur with radius of 1
Right:  Gaussian Blur with radius of 2

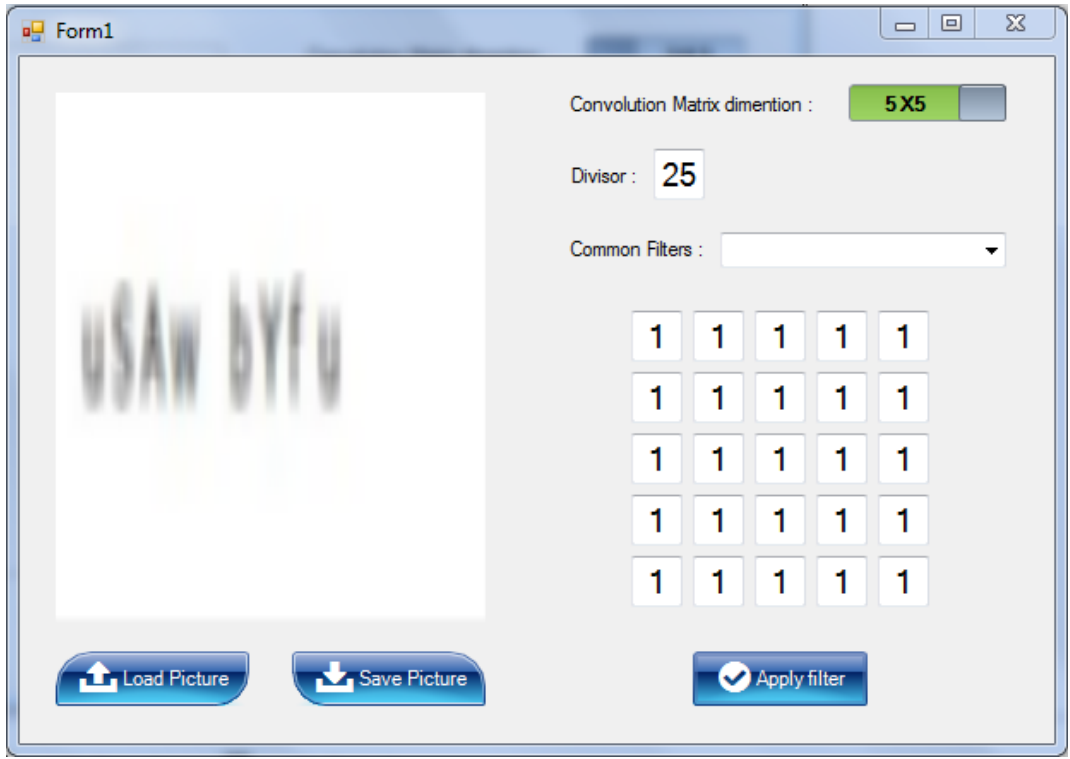

Figure 4. the application that was created to apply Gaussian filter to the images

To apply the Gaussian filter to an image in PHP (as a common webserver language) there are two ways, we can use "imagefilter" function [6] or we can implement the algorithm that has been explained above manually. To do so, we can use the "imagecolorat" function [7] to read each pixel's color and "imagesetpixel" function [8] to set the color of that pixel which obtained from applying the convolution matrix to it. The rest of the implementation is just loops and simple mathematical operations.

At first we have tested the readability of these two series of CAPTCHAs on human users by showing them the CAPTCHAs and asking them to enter what they saw. Also we asked them to rate the readability of these CAPTCHAs by giving a number between 1 and 10. We reached to satisfying results in this test. They were able to accurately distinguish 92% of the blurred CAPTCHAs with radius of 1 and 90% of them with radius of 2 and for readability rating we reached the average of 9.2 out of 10.

After that, we tested the security of CAPTCHAs against 2 OCR programs. These OCR programs were ReadIRIS 14 and Free OCR. In some cases they were unable to recognize text from background especially for the CAPTCHAs with radius of 2. In other cases mostly they were unable to recognize the exact words. You can see the complete results in Table 1 below.

Table 1. Results

| Humans Results | |
|---|---|
| Average characters similarity (radius of 1) | 99.11% |
| Average characters similarity (radius of 2) | 99.01% |
| Average exact match (radius of 1) | 92.00% |
| Average exact match (radius of 2) | 90.00% |
| Average readability rating provided by human testers | 9.2 |
| **OCR results** | |
| **ReadIRIS Application** | |
| Average characters similarity (radius of 1) | 31.86% |
| Average characters similarity (radius of 2) | 3.45% |
| Percentage of readable CAPTCHAs (radius of 1) | 56.00% |
| Percentage of readable CAPTCHAs (radius of 2) | 30.00% |
| Average exact match (radius of 1) | 16.00% |
| Average exact match (radius of 2) | 0.00% |
| **Free OCR Application** | |
| Average characters similarity (radius of 1) | 4.68% |
| Average characters similarity (radius of 2) | 0.52% |
| Percentage of readable CAPTCHAs (radius of 1) | 100.00% |
| Percentage of readable CAPTCHAs (radius of 2) | 32.00% |
| Average exact match (radius of 1) | 0.00% |
| Average exact match (radius of 2) | 0.00% |
| **Total Results** | |
| Total average characters similarity on OCR programs | 10.13% |
| Total average exact match on OCR programs | 4.00% |
| Total average characters similarity on humans | 99.06% |
| Total average exact match on humans | 91.00% |
| **Radius of 2 results** | |
| average exact match on OCR programs | 0.00% |
| average exact match on humans | 90.00% |

## 4. CONCLUSION

Because of some weaknesses of OCR programs, many webservers use OCR-Based CAPTCHAs to prevent futile registrations. In this paper we investigated the effect of using Gaussian filter on CAPTCHAs security. Actually we applied Gaussian blur filter on CAPTCHAs to improve their safety against OCR programs.

Based on the result we acquired (Table 1) the CAPTCHAs with Gaussian Blur filter applied on, are very powerful against OCR programs and also their readability by human users are extremely high. We generated two series of Blur CAPTCHAs, one with radius of one and the other with radius of two. Considering the results, blur CAPTHCAs with radius of two are more efficient than the other one. OCR programs couldn't recognize any of the CAPTCHAs with radius of two however human could recognize 90% of them. So they can be used in webservers to prevent abuse and unwanted registrations on them.

## ACKNOWLEDGMENT

We offer our thanks to the president and faculty members of Computer Science Department of Shahid Beheshti University.

Especially thanks to Young Researchers & Elite Club that provided the research possibility for me.